\documentclass[preprint,12pt]{elsarticle}

\usepackage{amssymb}

\usepackage{color}

\usepackage{dblfloatfix}

\journal{SPIE Journal of Medical Imaging}

\begin{document}

\begin{frontmatter}



\title{Deep Learning with Mixed Supervision for Brain Tumor Segmentation}



\author[a]{Pawel~Mlynarski }
\author[a]{Herv\'e~Delingette}
\author[b]{Antonio~Criminisi}
\author[a]{Nicholas~Ayache}
\address[a]{Universit\'e C\^ote d'Azur, Inria Sophia Antipolis, France.}
\address[b]{Microsoft Research Cambridge, United Kingdom.}

\begin{abstract}
Most of the current state-of-the-art methods for tumor segmentation are based on machine learning models trained on manually segmented images. This type of training data is particularly costly, as manual delineation of tumors is not only time-consuming but also requires medical expertise. On the other hand, images with a provided global label (indicating presence or absence of a tumor) are less informative but can be obtained at a substantially lower cost. In this paper, we propose to use both types of training data (fully-annotated and weakly-annotated) to train a deep learning model for segmentation. The idea of our approach is to extend segmentation networks with an additional branch performing image-level classification. The model is jointly trained for segmentation and classification tasks in order to exploit information contained in weakly-annotated images while preventing the network to learn features which are irrelevant for the segmentation task. We evaluate our method on the challenging task of brain tumor segmentation in Magnetic Resonance images from BRATS 2018 challenge. We show that the proposed approach provides a significant improvement of segmentation performance compared to the standard supervised learning. The observed improvement is proportional to the ratio between weakly-annotated and fully-annotated images available for training. 
\end{abstract}


\end{frontmatter}




\section{Introduction}
Cancer is today the third cause of mortality worldwide. In this paper, we focus on segmentation of gliomas, which are the most frequent primary brain cancers \cite{goodenberger2012genetics}. Gliomas are particularly malignant tumors and can be broadly classified according to their grade into low grade gliomas (grades I and II defined by World Health Organization) and high grades gliomas (grades III-IV). Glioblastoma multiforme is the most malignant form of glioma and is associated with a very poor prognosis: the average survival time under therapy is between 12 and 14 months.

Medical images play a key role in diagnosis, therapy planning and monitoring of cancers. Treatment protocols often include evaluation of tumor volumes and locations. In particular, for radiotherapy planning, clinicians have to manually delineate target volumes, which is a difficult and time-consuming task. Magnetic Resonance (MR) images \cite{bauer2013survey} are particularly suitable for brain cancer imaging. Different MR sequences (T2, T2-FLAIR, T1, T1+gadolinium) highlight different tumor subcomponents such as edema, necrosis or contrast-enhancing core.

In recent years, machine learning methods have achieved impressive performance in a large variety of image recognition tasks. Most of the recent state-of-the-art segmentation methods are based on Convolutional Neural Networks (CNN) \cite{lecun1995convolutional, long2015fully}. CNNs have the considerable advantage of automatically learning relevant image features. This ability is particularly important for the tumor segmentation task. CNN-based methods \cite{pereira2015deep, kamnitsas2016efficient, kamnitsas2017ensembles, wang2017automatic} have obtained the best performances on the four last editions of Multimodal Brain Tumor Segmentation Challenge (BRATS) \cite{menze2015multimodal,bakas2017advancing}.

Most of the segmentation methods based on machine learning rely uniquely on manually segmented images. The cost of this annotation is particularly high in medical imaging where manual segmentation is not only time-consuming but also requires high medical competences. Image intensity of cancerous tissues in MRI or CT scans is often similar to the one of surrounding healthy or pathological tissues, making the exact tumor delineation difficult and subjective. In the case of brain tumors, according to \cite{menze2015multimodal}, the inter-rater overlap of expert segmentations is between 0.74 and 0.85 in terms of Dice coefficient. For these reasons, high-quality manual tumor segmentations are generally available in very limited numbers. Segmentation approaches able to exploit images with weaker forms of annotations are therefore of particular interest.

\begin{figure*}[!h]
\centering
\includegraphics[width=1.0\textwidth]{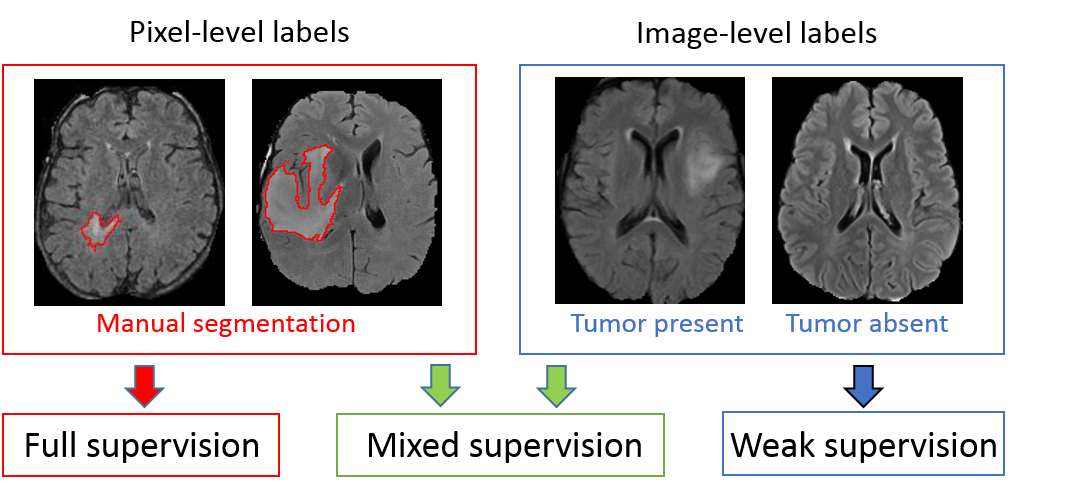}
\caption{Different levels of supervision for training of segmentation models. Standard models are trained on fully-annotated images only, with pixel-level labels. Weakly-supervised approaches aim to train models using only weakly-annotated images, e.g. with image-level labels. Our model is trained with a mixed supervision, exploiting both types of training images.}
\label{fig_intro}
\end{figure*}

In this paper, we assume that the training dataset contains two types of images: fully-annotated (with provided ground truth segmentation) and weakly-annotated, with an image-level label indicating presence or absence of a tumor tissue within the image (Fig. \ref{fig_intro}). We refer to this setting as 'mixed supervision'. The latter type of annotations can be obtained at a substantially lower cost as it is less time-consuming, potentially requires less medical expertise and can be obtained without the use of a dedicated software.

We introduce a novel CNN-based segmentation model which can be trained using weakly-annotated images in addition to fully-annotated images. We propose to extend segmentation networks, such as U-Net \cite{ronneberger2015u}, with an additional branch, performing image-level classification. The model is trained jointly for both tasks, on fully-annotated and weakly-annotated images. The goal is to exploit the representation learning ability of CNNs to learn from weakly-annotated images while supervising the training using fully-annotated images in order to learn features relevant for the segmentation task. Our approach differs from the standard semi-supervised learning as we consider weakly-annotated data instead of totally unlabelled data. To the best of our knowledge, we are the first to combine pixel-level and image-level labels for training of models for tumor segmentation.

We perform a series of cross-validated tests on the challenging task of segmentation of gliomas in MR images from BRATS 2018 challenge. We evaluate our model both for binary and multiclass segmentation using a variable number of ground truth segmentations available for training. Since all 3D images from the BRATS 2018 contain brain tumors, we focus on the 2D problem of tumor segmentation in axial slices of a MRI and we assume slice-level labels for weakly-annotated images. Using approximately 220 MRI with slice-level labels and a varying number of fully-annotated MRI, we show that our approach significantly improves the segmentation accuracy when the number of fully-annotated cases is limited.

\section{Related work}
In the literature, there are several works related to weakly-supervised and semi-supervised learning for object segmentation or detection. Most of the related works were applied to natural images. 

The first group of weakly-supervised methods aims to localize objects using only weakly-annotated images for training. When only image-level labels are available, one approach is to design a neural network which outputs two feature maps per class (interpreted as 'heat maps' of the class) which are then pooled to obtain an image-level classification score penalized during the training \cite{pathak2014fully,pinheiro2015image,saleh2016built,bearman2016s,wang2018automated}. At test time, these 'heat maps' are used for detection (determining a bounding box of the object) or segmentation. To guide the training process, some works use self-generated spatial priors \cite{pinheiro2015image,saleh2016built,bearman2016s} or inconsistency measures \cite{wang2018automated} in the loss function. To obtain an image-level score, in \cite{pathak2014fully,bearman2016s}, global maximum pooling is used. Application of the maximum function on large feature maps may cause optimization problems as training of neural networks is based on computation of gradients \cite{dreyfus1990artificial}. LogSumExp approximation of the maximum \cite{boyd2004convex} is therefore used in the works \cite{pinheiro2015image,saleh2016built} in order to partially limit this problem. Average pooling on small feature maps was used by Wang et al \cite{wang2018automated} for the problem of detection of lung nodules.

Another type of weakly-supervised methods aims to detect objects in natural images based on classification of image subregions \cite{girshick2014rich, oquab2015object} using pre-trained classification networks such as VGG-Net \cite{simonyan2014very} or AlexNet \cite{krizhevsky2012imagenet}. In fact, one particularity of natural images is their recursive aspect: one image can correspond to a subpart of another image (e.g. two images of the same object taken from different distances). A classification network trained on a large dataset may therefore be used on a subregion of a new image in order to determine if it contains an object of interest.

Pre-trained classification networks were also used to detect objects by determining image subregions whose modification influences the global classification score of a class. In \cite{simonyan2013deep}, Simonyan et al. propose to compute the gradient of the classification score with respect to the intensities of pixels and to threshold it in order to localize the object of interest. However, these partial derivatives represent a very weak information for tumor segmentation, which requires a complex analysis of the spatial context. The method proposed in \cite{bergamo2014self} is based on replacing image subregions by the mean value in order to measure the drop of the classification score.

Overall, the reported segmentation performances of weakly-supervised methods are considerably lower than the ones obtained by semi-supervised and supervised approaches. In absence of pixel-level labels, a model may learn irrelevant features, due for example to co-occurrences of objects or image acquisition differences in the case of multicenter medical data. Despite the cost of manual segmentation, at least few fully-annotated images can still be obtained in many cases. 

In standard semi-supervised learning \cite{cheplygina2018not} for classification, the training data is composed both of labelled samples and unlabelled samples. Unlabelled samples can be used to encourage the model to satisfy some properties on relations between labels and the feature space. Common properties include smoothness (points close in the feature space should be close in the target space), clustering (labels form clusters in the feature space) and low density separation (decision boundaries should be in low density regions of the feature space). Semi-supervised learning based on these properties can be performed by graph-based methods such as the recent work of Kamnitsas et al. \cite{kamnitsas2018semi}. The main idea of such methods is to propagate labels in a fully-connected graph whose nodes are samples (labelled and unlabelled) and whose edges are weighted by similarities between samples. The use of graph-based semi-supervised methods is difficult for segmentation, in particular because it implies computation of similarity metrics between samples, whereas each single image is generally composed of millions of samples (pixels or voxels).

Relatively few works were proposed for semi-supervised learning for image segmentation. Some semi-supervised approaches are based on self-training, i.e. training of a machine learning model on self-generated labels. Iterative algorithms similar to EM \cite{zhang2001segmentation} were proposed for natural images \cite{papandreou2015weakly} and medical images \cite{rajchl2016deepcut}. Recently, Hung et al \cite{hung2018adversarial} proposed a method based on Generative Adversarial Networks \cite{goodfellow2014generative} where the generator network performs image segmentation and the discriminator network tries to determine if a segmentation corresponds to the ground truth or the segmentation produced by the generator. The discriminator network is used to produce confidence maps for self-training. The approaches based on self-training have the drawback of learning on uncertain labels (produced by the model itself) and training of such models is difficult.

Other approaches assume mixed levels of supervision similarly to our approach. Hong et al \cite{hong2015decoupled,hong2016learning} proposed decoupled classification and segmentation, an approach for segmentation of objects in natural images based on a two-step training with a varying level of supervision. This architecture is composed of two separate networks trained sequentially, one performing image-level classification and used as encoder, and the another one taking as input small feature maps extracted from the encoder and performing segmentation. An important drawback of such design, in the case of tumor segmentation, is that the segmentation network does not take as input the original image and can therefore miss important details of the image (e.g. small tumors).

Our approach is related to multi-task learning \cite{evgeniou2004regularized}. In our case, the goal of training for two tasks (segmentation and classification) is to exploit all the available labels and to guide the training process to learn relevant features. The approach closest to ours is the one of Shah et al. \cite{shah2018ms}. In this work, the authors consider three types of annotations: segmentations, bounding boxes and seed points at the borders of objects. A neural network is trained using these three types of training data. In our work, we exploit the use of a significantly weaker form of annotations, image-level labels.

\section{Joint classification and segmentation with Convolutional Neural Networks}
\subsection{Deep learning model for binary segmentation}
We designed a novel deep learning model, which aims to take advantage of all available voxelwise and image-level annotations. We propose to extend a segmentation CNN with an additional subnetwork performing image-level classification and to train the model for the two tasks jointly. Most of the layers are shared between the classification and segmentation subnetworks in order to transfer the information between the two subnetworks. In this paper we present the 2D version of our model, which can be used on different types of medical images such as slices of a CT scan or a multisequence MRI.

The proposed network takes as input an image of dimensions 300x300 and extends U-net \cite{ronneberger2015u} which is currently one of the most used architectures for segmentation tasks in medical imaging. This segmentation architecture is composed of an encoder part and a decoder part which are connected by concatenations between layers at the same scale, in order to combine low-level and local features with high-level and global features. This design is well suited for the tumor segmentation task since the classification of a voxel as tumor requires to compare its value with its close neighborhood but also taking into account a large spatial context. The last convolutional layer of U-net produces pixelwise classification scores, which are normalized by softmax function during the training phase. We apply batch normalization \cite{ioffe2015batch} in all convolutional layers except the final layer. 

\begin{figure*}[!b]
\centering
\includegraphics[width=1.0\textwidth]{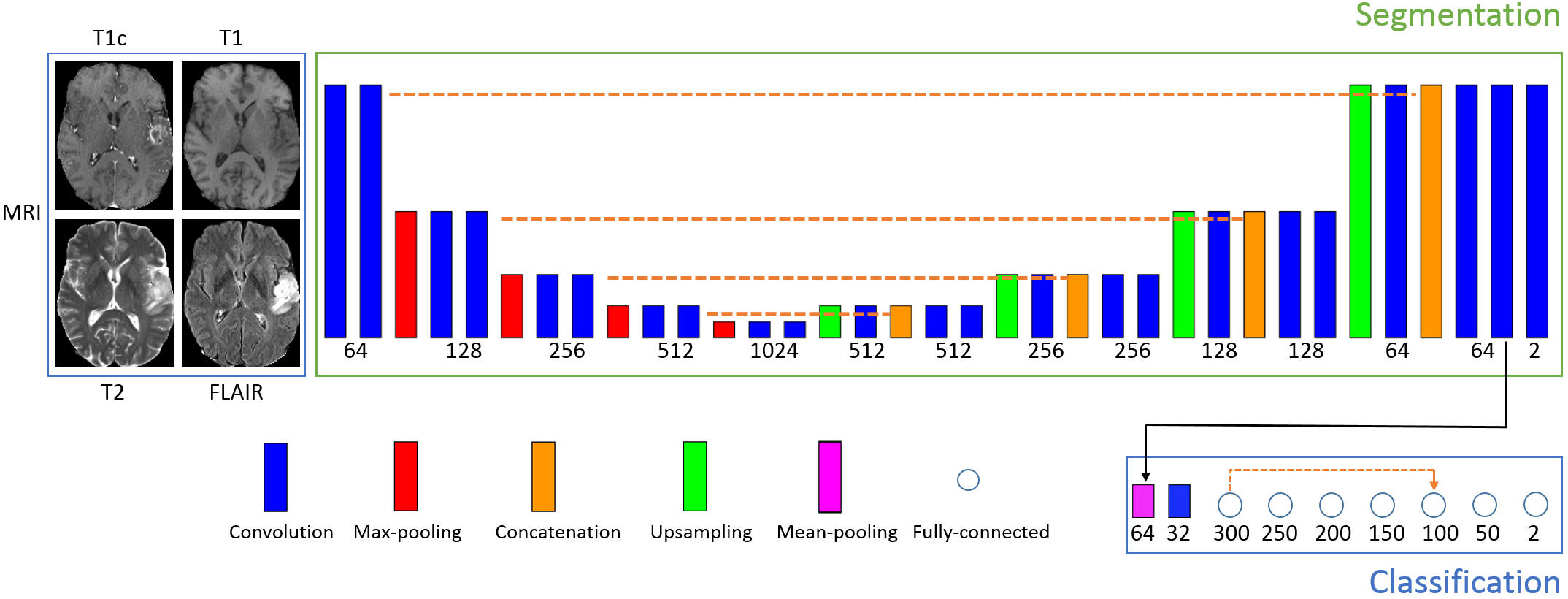}
\caption{Architecture of our model for binary segmentation. The numbers of outputs are specified below boxes representing layers. The height of rectangles represents the scale (increasing with pooling operations). The dashed lines represent concatenation operations. The proposed architecture is an extended version of U-net, with a subnetwork performing image-level classification. Training of the model corresponds to a joint minimization of two loss functions related respectively to segmentation and image-level classifcation tasks.}
\label{fig_model_binary}
\end{figure*}

We propose to add an additional branch to the network, performing image-level classification (Fig. \ref{fig_model_binary}), in order to exploit the information contained in weakly-annotated images during the training. This classification branch takes as input the second to last convolutional layer of U-net (representing a rich information extracted from a local and a long-range spatial context) and is composed of one mean-pooling, one convolutional layer and 7 fully-connected layers.

The goals of taking a layer from the final part of U-Net as input of the classification branch are both to guide the image-level classification task and to force the major part of the segmentation network to take into account weakly-annotated images. This also helps the optimization process by taking advantage of the connectivity of layers in U-Net, helping the flow of gradients of the loss function during the training (in particular, note the connection between the first part and the last part of U-Net). 

The second to last layer of the segmentation network outputs 64 feature maps of size 101x101 from which the classification branch has to output two global (image-level) classification scores (tumor absent/tumor present). We first reduce the size of these feature maps by applying a mean-pooling with kernels of size 8x8 and the stride of 8x8. We use the mean pooling rather than max-pooling in order to avoid information loss and optimization problems. One convolutional layer, with ReLU activation and kernels of size 3x3, is then added to reduce the number of feature maps from 64 to 32. The resulting 32 feature maps of size 11x11 are the input of the first fully-connected layer of the classification branch.

According to our experiments, a relatively deep architecture of the classification branch with a limited number of parameters and a skip-connection between layers yields the best performance. This observation is in agreement with current common designs of neural networks. Deep networks have the capacity to learn more complex features, due to applied non-linearities. The connectivity between layers at different depths helps the optimization process (e.g. Res-Net \cite{he2016deep}). In our case, we use 7 fully-connected layers with ReLU activations (except the final layer) and we concatenate the outputs of the first and the fifth fully-connected layer. The last fully-connected layer outputs image-level classification scores (tumor tissue absent or present).

The model is trained both on fully-annotated and weakly-annotated images for the two tasks jointly (segmentation and classification). We can distinguish between three types of training images. First, images containing a tumor and with provided ground truth segmentation are the most costly ones. The second type are images that do not contain tumor, which implies that none of their pixels corresponds to a tumor. In this case, the ground truth segmentation is simply the zero matrix. The only problematic case is the third one, when the image is labelled as containing a tumor but without provided segmentation. 

To train our model, we propose to form training batches containing the three mentioned types of images: $k$ positive cases (containing a tumor) with provided segmentation, $m$ negative cases and $n$ positive cases without provided segmentation. In our experiments we chose $k=4$, $m=2$, $n=4$.



Given a training batch $b$ and the network parameters $\theta$, we use a weighted pixelwise cross-entropy loss on images of types 1 and 2: 
\begin{equation}
Loss_{s}^b (\theta)= -\frac{1}{P} \sum_{ i=1}^{k+m} \sum_{ (x,y)} w^i_{ (x,y)} \log(p^l_{i, (x,y)}(\theta))
\end{equation}
where P is the number of pixels; $p^l_{i, (x,y)}$ is the classification score given by the network to the ground truth label for pixel (x,y) of the $i^{th}$ image of the batch and $w^i_{ (x,y)}$ is the weight given to this pixel. The weights are used to limit the effect of class imbalance, since tumor pixels represent a small portion of the image. Weights of pixels are set automatically according to the composition of the training batch (number of pixels of each class) so that pixels associated with healthy tissues have a total weight of $t_0$ in the loss function and the pixels of the tumor class have a total weight of $t_1$, where $t_0$ and $t_1$ are target weights fixed manually ($t_0=0.7$ and $t_1=0.3$ in our experiments). It means that if the training batch contains $N_t$ pixels labelled as tumor, then each tumor pixel has a weight of $t_1/N_1$ (the pixelwise weight is high when the number of tumor pixels is low). We fix a higher target weight for healthy tissues to avoid oversegmentation, while increasing the relative weight of tumor pixels ($t_1=0.3$) compared to the standard non-weighted cross-entropy loss where the relative weight of the tumor class is proportional to the number of tumor pixels, i.e approximately 1-2 \%. This type of loss function was used in our previous work \cite{mlynarski20183d}.

The classification loss  is a standard cross-entropy loss on all images of the training batch: $Loss_{c}= -\frac{1}{k+m+n} \sum_{ i=1}^{k+m+n} \log(p^l_{i} (\theta))$ where $p^l_{i}$ is the global classification score given by the network to the ground truth global label for the $i^{th}$ image of the batch. In particular, fully-annotated images are also used for training of the classification branch in order to transfer the knowledge from the segmentation task to the image-level classification. We do not apply weights on the classification loss as image-level labels are balanced through the sampling of training batches (having a fixed number of non-tumor images).

Since both segmentation and classification losses are normalized, we define the total loss as a convex combination of the classification and segmentation losses: {$Loss= a* Loss_{s} + (1-a)* Loss_{c}$}. In our experiments we fixed a=0.7, i.e. we give a higher importance to segmentation errors. The determination of this parameter was performed empirically through a series of cross-validated tests.

We train our model with a variant of Stochastic Gradient Descent (SGD) with momentum \cite{rumelhart1988learning}, used also in our previous work \cite{mlynarski20183d}. The main differences with the standard SGD are to divide the gradient by its norm and to compute gradients on several training batches in each iteration, in order to take into account many training examples while bypassing GPU memory constraints.

\subsection{Extension to the multiclass problem}

We extend our model to the multiclass case where each pixel has to be labelled with one of K classes, such as the four ones considered in BRATS challenge (non-tumor, contrast-enhancing core, edema, non-enhancing core). We now assume that image-level labels are provided for each class (absent/present in the image).

Extension of the segmentation subnetwork to the multiclass problem is straightforward, by changing the number of final feature maps to match the number of classes. However, image-level labels are not exclusive, i.e. an image may contain several tumor subclasses. For this reason, we propose to consider one image-level classification output per tumor subclass, indicating absence or presence of the given subclass.

\begin{figure*}[t!]
\centering
\includegraphics[width=1.0\textwidth]{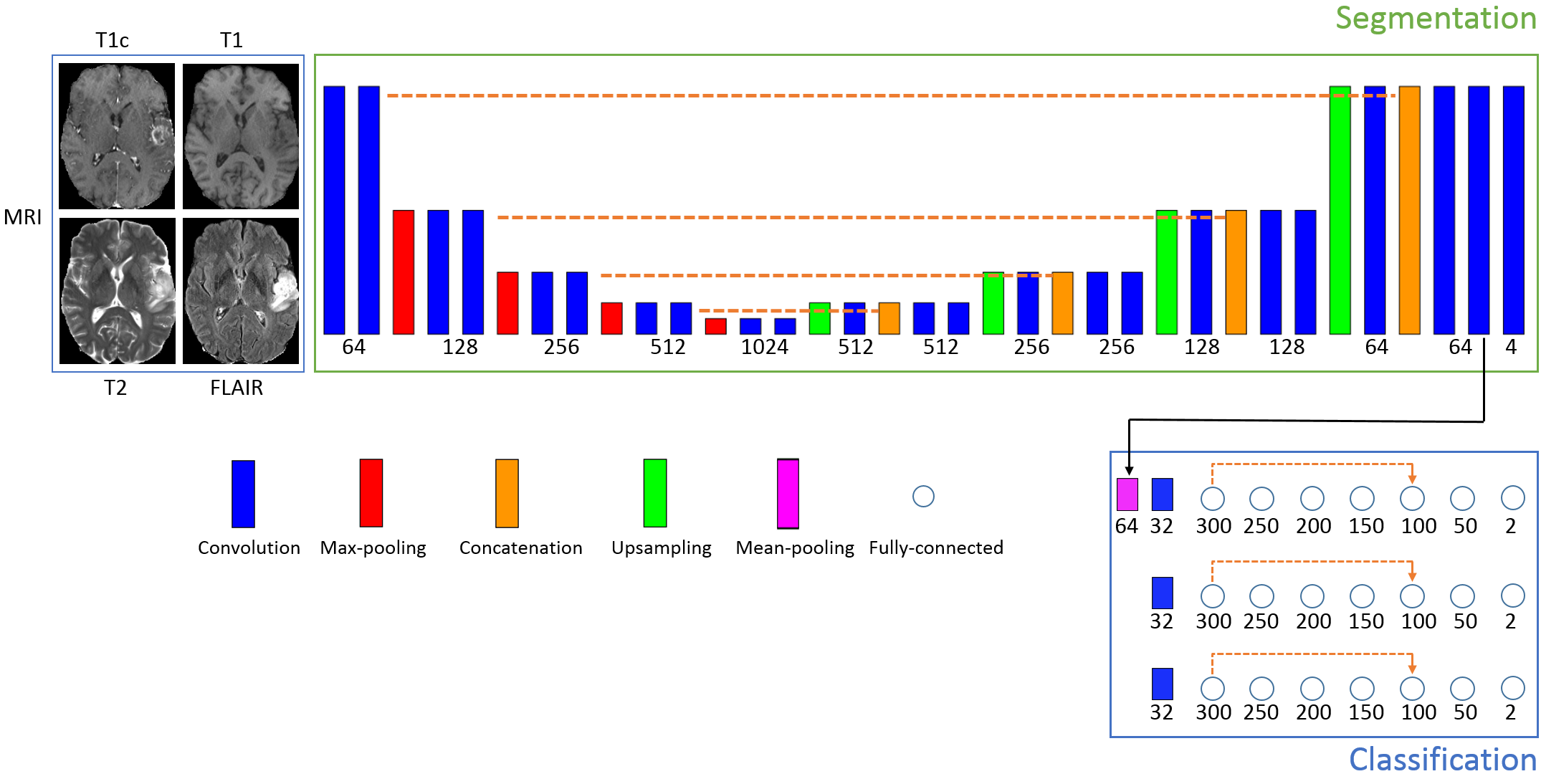}
\caption{Extension of our model to the multiclass problem. The number of final feature maps of the segmentation subnetwork is equal to the number of classes (4 in our case). As image-level labels (class present/absent) are not exclusive, we consider one classification branch per tumor subclass.}
\label{fig_model_multiclass}
\end{figure*}

According to our experiments, better performances are obtained when each subclass has its dedicated entire classification branch (Fig. \ref{fig_model_multiclass}). A possible reason is that the image-level classification of tumor subclasses is a challenging task requiring a sufficient number of dedicated parameters.

Training batches are sampled similarly to the binary case, however each tumor subclass has to be present at least once in each training batch. In our implementation, we store lists of paths of images containing tumor subclasses in order to sample from these lists during the training of the model.


In the segmentation loss we empirically fix the following target weights for the four classes (non-tumor, non-enhancing tumor core, edema, enhancing-core): $t_0=0.7$, $t_1=0.1$,  $t_2=0.1$,  $t_3=0.1$ (all tumor subclasses have an equal weight in the loss function). The loss associated with each classification branch is the same as in the binary case and the total classification loss is the average across all classification branches. We observe the need to decrease the parameter $a$ (weight of the segmentation loss) compared to the binary case. We {fixed $a=0.3$} according to performed cross-validated experiments.

\section{Experiments}
\subsection{Data}

We evaluate our method on the challenging task of brain tumor segmentation in multisequence MR scans, using the \textit{Training} dataset of BRATS 2018 challenge. It contains 285 multisequence MRI of patients diagnosed with low-grade gliomas or high-grade gliomas. For each patient, manual ground truth segmentation is provided. In each case, four MR sequences are available: T1, T1+gadolinium, T2 and FLAIR (Fluid Attenuated Inversion Recovery). Preprocessing performed by the organizers includes skull-stripping, resampling to 1 $mm^3$ resolution and registration of images to a common brain atlas. The resulting volumes are of size 240x240x155. The images were acquired in 19 different imaging centers. In order to normalize image intensites, each image is divided by the median of non-zero voxels  (which is supposed to be less affected by the tumor zone than the mean) and multiplied the image by a fixed constant.

Each voxel is labelled with one of the following classes: non-tumor (class 0), contrast-enhancing core (class 3), non-enhancing core (class 1), edema (class 2). The benchmark of the challenge groups classes in three regions: \textit{whole tumor} (all tumor subclasses), \textit{tumor core} (classes 1 and 3, corresponding to the visible tumor mass) and \textit{enhancing core} (class 3).

\begin{figure}[t!]
\includegraphics[width=0.9\textwidth]{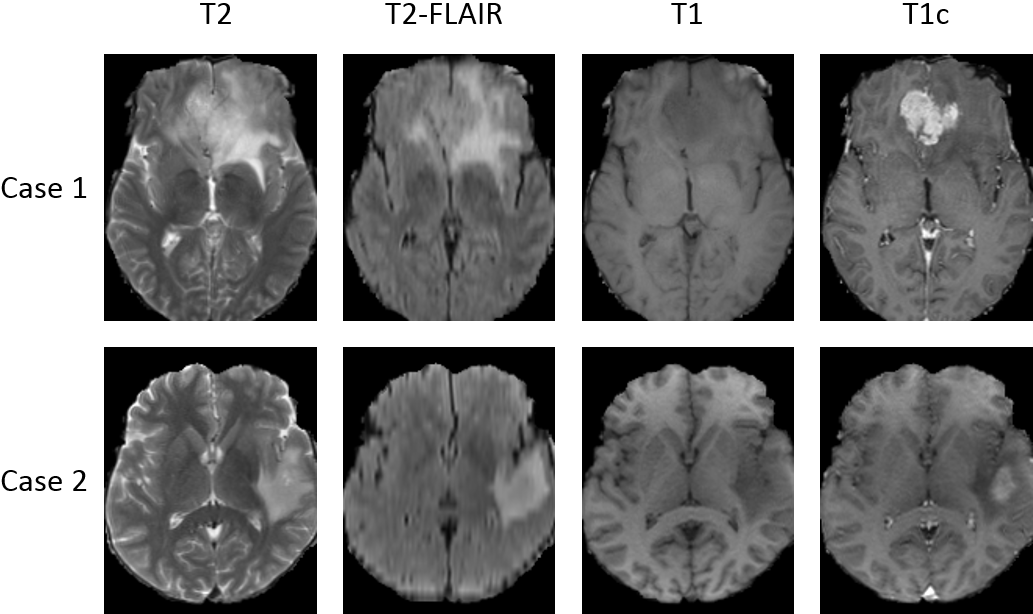}
\caption{Examples of multisequence MRI from the BRATS 2018 database. While T2 and T2-FLAIR highlight the edema induced by the tumor, T1 is suitable for determining the tumor core. In particular, T1 acquired after injection of a contrast product (T1c) highlights the tumor angiogenesis, indicating presence of highly proliferative cancer cells.}
\label{fig_images}
\end{figure}

Given that all 3D images of the database contain tumors (no negative cases to train a 3D classification network), we consider the 2D problem of tumor segmentation in axial slices of the brain.

\subsection{Test setting}
The goal of our experiments is to compare our approach with the standard supervised learning.  In each of the performed tests, our model is trained on fully-annotated and weakly-annotated images and is compared with the standard U-Net trained on fully-annotated images only. The goal is to compare our model with a commonly used segmentation model on a publicly available database.

We consider three different training scenarios, with a varying number of patients for which we assume a provided manual tumor segmentation. In each scenario we perform a 5-fold cross-validation. In each fold, 57 patients are used for test and 228 patients are used for training. Among the 228 training images, few cases are assumed to be fully-annotated and the remaining ones are considered to be weakly-annotated, with slice-level labels. The fully-annotated images are different in each fold. If the 3D volumes are numbered from 0 to 284, then in $k^{th}$ fold, the test images correspond to the interval [(k-1)*57, k*57 -1], the next few images correspond to fully-annotated images and the remaining ones are considered as weakly-annotated (the folds are generated in a circular way). In the following, \textit{FA} denotes the number of fully-annotated cases and \textit{WA} denotes the number of weakly-annotated cases (with slice-level labels).

In the first training scenario, 5 patients are assumed to be provided with a manual segmentation and 223 patients have slice-level labels. In the second and the third scenario, the numbers of fully-annotated cases are respectively 15 and 30 and the numbers of weakly-annotated images are therefore respectively 213 and 198. The three training scenarios are independent, i.e. folds are re-generated randomly (the list of all images is permuted randomly and the folds are generated). In fact, results are likely to depend not only on the number of fully-annotated images but also on qualitative factors (for example the few fully-annotated images may correspond to atypical cases), and the goal is to test the method in various settings. Overall, our approach is compared to the standard supervised learning on 60 tests (5-fold cross-validation, three independent training scenarios, three binary problems and one multiclass problem). 

We evaluate our method both on binary segmentations problems (separately for each of three tumor regions considered in the challenge) and on the end-to-end multiclass segmentation problem. In each binary case, the model is trained for segmentation and classification of one tumor region (whole tumor, tumor core or enhancing core).

Segmentation performance is expressed in terms of Dice score quantifying the overlap between the ground truth ($Y$)  and the output of a model ($\tilde{Y}$):
\begin{equation}
 DSC(\tilde{Y}, Y)= \frac{2 |\tilde{Y} \cap Y|}{|\tilde{Y}|+|Y|}
\end{equation}

\subsection{Results}
The main observation is that our model with mixed supervision provides a significant improvement over the standard supervised approach (U-Net trained on fully-annotated images) when the number of fully-annotated images is limited. In the two first training scenarios (5 FA and 15 FA), our model outperformed the supervised approach on the three binary segmentation problems (Table \ref{tab_results1}) and in the multiclass setting (Table \ref{tab_results2}). The largest improvements are in the first scenario (5 FA) for the \textit{whole tumor} region where the improvement is of 8 points of the mean Dice score in the binary setting and of 9 points of Dice in the multiclass setting. Results on different folds of the second scenario (intermediate case, 15 FA) are displayed in Table \ref{tab_folds1} for the binary problems and in Table \ref{tab_folds2} for the multiclass problem. Our approach provided an improvement in all folds of the second scenario and for all tumor regions, except one fold for \textit{enhancing core} in the binary setting. In the third scenario (30 FA + 198 WA), our approach and the standard supervised approach obtained similar performances. 

\begin{table}[t!]

\centering
\caption{Mean Dice scores (5-fold cross-validation, 57 test cases in each fold) in the three binary segmentation problems obtained by the standard supervised approach and by our model trained with mixed supervision.}
\begin{tabular}{|c|c|c|c|}
  \hline
& \scriptsize{Whole Tumor}& \scriptsize{Tumor Core}& \scriptsize{Enhancing core}\\ 
\hline
\scriptsize{Standard supervision 5 FA}& \scriptsize{70.39 }& \scriptsize{48.14}& \scriptsize{55.74  }\\ 
\hline
\scriptsize{Mixed supervision 5 FA + 223 WA}& \scriptsize{\textbf{78.34}   }& \scriptsize{\textbf{50.11}}& \scriptsize{\textbf{60.06}  }\\ 
\hline
\hline
\scriptsize{Standard supervision 15 FA}& \scriptsize{77.91   }& \scriptsize{58.33}& \scriptsize{62.88  }\\ 
\hline
\scriptsize{Mixed supervision 15 FA + 213 WA}& \scriptsize{\textbf{80.92}   }& \scriptsize{\textbf{63.23}}& \scriptsize{\textbf{66.61}  }\\ 
\hline
\hline
\scriptsize{Standard supervision 30 FA}& \scriptsize{\textbf{83.95}   }& \scriptsize{66.17}& \scriptsize{\textbf{69.15}  }\\ 
\hline
\scriptsize{Mixed supervision 30 FA + 198 WA}& \scriptsize{83.84   }& \scriptsize{\textbf{68.30}}& \scriptsize{67.18  }\\ 
\hline
\hline
\scriptsize{Standard supervision 228 FA}& \scriptsize{86.80}& \scriptsize{77.09}& \scriptsize{72.20  }\\ 
\hline

  \hline
\end{tabular}
\label{tab_results1}
\end{table}

\begin{table}[t!]

\centering
\caption{Results obtained for the three binary problems (whole tumor, tumor core, enhancing core) on different folds in the case with 15 fully-annotated images and 213 weakly-annotated images.}
\begin{tabular}{|c|c|c|c|c|c|c|}
  \hline
 & \scriptsize{fold1}& \scriptsize{fold2}& \scriptsize{fold3}& \scriptsize{fold4}& \scriptsize{fold5}& \scriptsize{mean}\\
\hline
\scriptsize{Standard supervision, whole tumor} &\scriptsize{76.23}&\scriptsize{78.15}&\scriptsize{78.13}&\scriptsize{77.67}&\scriptsize{79.35}&\scriptsize{77.91}\\ 
\hline
\scriptsize{Mixed supervision, whole tumor}&\scriptsize{\textbf{82.36}}&\scriptsize{\textbf{81.03}}&\scriptsize{\textbf{78.96}}&\scriptsize{\textbf{79.88}}&\scriptsize{\textbf{82.35}}&\scriptsize{\textbf{80.92}}\\ 
\hline
\hline
\hline
\scriptsize{Standard supervision, tumor core}&\scriptsize{61.46}&\scriptsize{61.17}&\scriptsize{56.68}&\scriptsize{56.42}&\scriptsize{55.94}&\scriptsize{58.33}\\ 
\hline
\scriptsize{Mixed supervision, tumor core}& \scriptsize{\textbf{63.15}}&\scriptsize{\textbf{66.82}}&\scriptsize{\textbf{63.45}}&\scriptsize{\textbf{60.83}}&\scriptsize{\textbf{61.91}}&\scriptsize{\textbf{63.23}}\\ 
\hline
\hline
\hline
\scriptsize{Standard supervision, enhancing core}&\scriptsize{66.33}&\scriptsize{61.08}&\scriptsize{57.86}&\scriptsize{\textbf{68.09}}&\scriptsize{61.02}&\scriptsize{62.88}\\ 
\hline
\scriptsize{Mixed supervision, enhancing core}& \scriptsize{\textbf{68.72}}&\scriptsize{\textbf{70.65}}&\scriptsize{\textbf{60.34}}&\scriptsize{67.55}&\scriptsize{\textbf{65.80}}&\scriptsize{\textbf{66.61}}\\ 
\hline

  \hline
\end{tabular}
\label{tab_folds1}
\end{table}

\begin{figure}[ht!]
\includegraphics[width=1.0\textwidth]{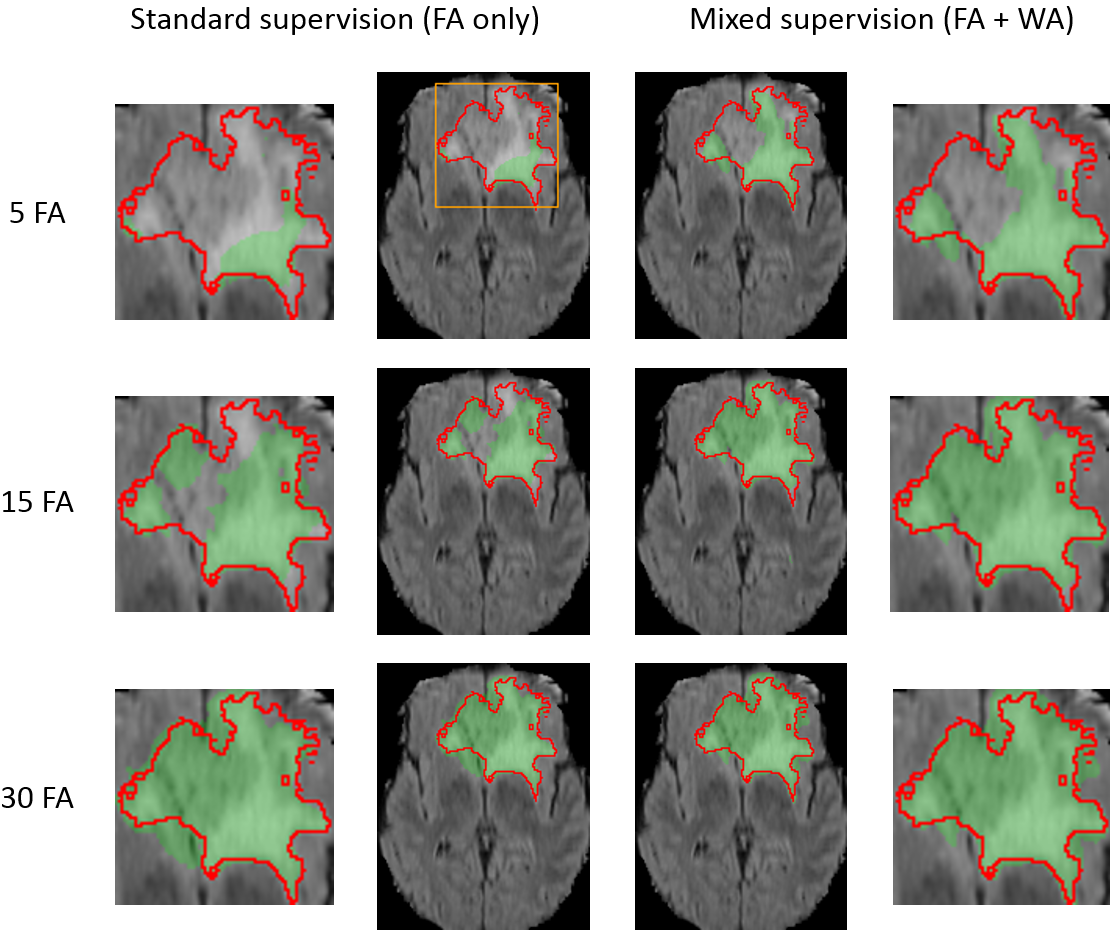}
\caption{Comparison of our approach with the standard supervised learning for binary segmentation of the 'whole tumor' region. Each row represents the same test example (first image of Fig. 4) from a different training scenario (5, 15 or 30 fully-annotated scans available for training). FA and WA refer respectively to the number of fully-annotated MRI and weakly-annotated MRI (with slice-level labels). The results are displayed on MRI T2-FLAIR sequence. The performance of both models improves with the number of manual segmentations available for training.}
\label{fig_results_whole_tumor}
\end{figure}

\begin{figure}[ht!]
\includegraphics[width=1.0\textwidth]{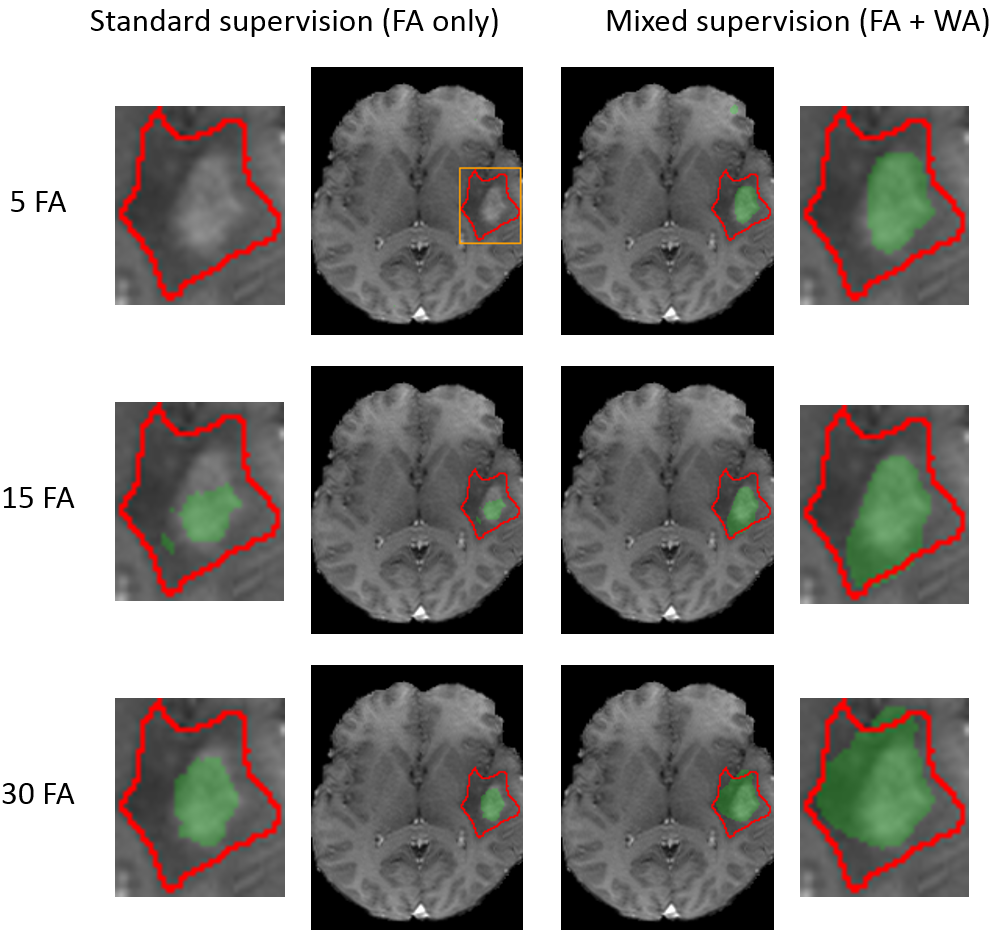}
\caption{Comparison of our approach with the standard supervised learning for binary segmentation of the 'tumor core' region (test example corresponding to the bottom image of Fig. 4). Each row corresponds to a different training scenario (5, 15 or 30 fully-annotated scans available for training). FA and WA refer to the numbers of fully-annotated and weakly-annotated scans. The results are displayed on MRI T1+gadolinium. The observations are similar to the problem of binary segmentation of the 'whole tumor' region. In particular, in the first training scenario, the standard supervised approach does not detect the tumor core zone, in contrast to our method.}
\label{fig_results_tumor_core}
\end{figure}

\begin{figure}[ht!]
\includegraphics[width=0.75\textwidth]{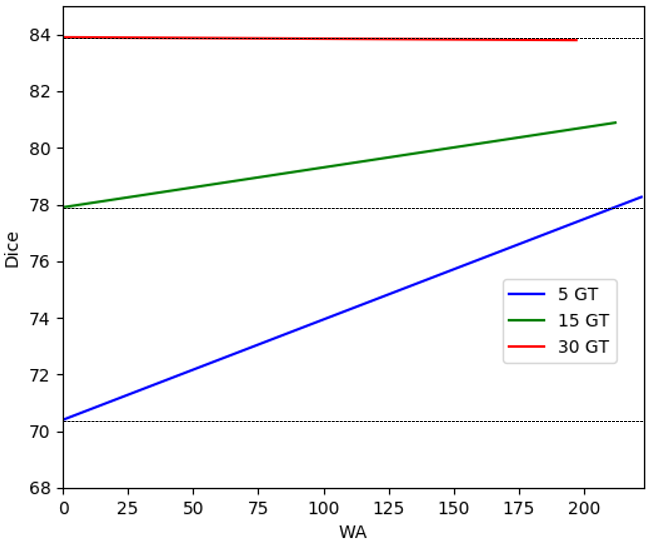}
\caption{Illustration of the improvement provided by the mixed supervision for binary segmentation of the 'whole tumor' region. Mixed supervision using 5 fully-annotated MRI and 223 weakly-annotated MRI obtains a slightly better performance than the standard supervised approach using 15 fully-annotated MRI. The improvement provided by the weakly-annotated images decreases with the number of available ground truth segmentations.}
\label{fig_curve_whole_tumor}
\end{figure}

Segmentation performance increases quickly with the first fully-annotated cases, both for the standard supervised learning and the learning with mixed supervision. For instance, mean Dice score obtained by the supervised approach for \textit{whole tumor} increases from 70.39, in the case with 5 fully-annotated images, to 77.9 in the case with 15 fully-annotated images. Our approach using 5 fully-annotated images and 223 weakly-annotated images obtained a slightly better performance (78.3) than the supervised approach using 15 fully-annotated cases (77.9). This result is represented on Fig. \ref{fig_curve_whole_tumor}.

Note that each fully-annotated case corresponds to a large 3D volume with voxelwise annotations. Each manually segmented axial slice of size 240x240 corresponds to 57 600 labels, which represents indeed a huge amount of information compared to one global label simply indicating presence of absence of a tumor tissue within the slice.

\begin{table}[b!]

\centering
\caption{Mean Dice scores (5-fold cross-validation, 57 test cases in each fold) obtained by the standard supervised approach and by our model in the multiclass setting.}
\begin{tabular}{|c|c|c|c|}
  \hline
& \scriptsize{Whole Tumor}& \scriptsize{Tumor Core}& \scriptsize{Enhancing core}\\ 
\hline
\scriptsize{Standard supervision 5 FA}& \scriptsize{67.61 }& \scriptsize{51.12}& \scriptsize{58.15  }\\ 
\hline
\scriptsize{Mixed supervision 5 FA + 223 WA}& \scriptsize{\textbf{76.64}   }& \scriptsize{\textbf{56.30}}& \scriptsize{\textbf{58.19}  }\\ 
\hline
\hline
\scriptsize{Standard supervision 15 FA}& \scriptsize{74.46   }& \scriptsize{59.87}& \scriptsize{61.85  }\\ 
\hline
\scriptsize{Mixed supervision 15 FA + 213 WA}& \scriptsize{\textbf{79.39}   }& \scriptsize{\textbf{63.91}}& \scriptsize{\textbf{65.71}  }\\ 
\hline
\hline
\scriptsize{Standard supervision 30 FA}& \scriptsize{81.10   }& \scriptsize{\textbf{67.48}}& \scriptsize{\textbf{68.67}  }\\ 
\hline
\scriptsize{Mixed supervision 30 FA + 198 WA}& \scriptsize{\textbf{81.23}   }& \scriptsize{66.33}& \scriptsize{67.69  }\\ 
\hline
\hline
\scriptsize{Standard supervision 228 FA}& \scriptsize{85.67  }& \scriptsize{78.78}& \scriptsize{74.14 }\\ 
\hline

  \hline
\end{tabular}
\label{tab_results2}
\end{table}

\begin{table}[t!]

\centering
\caption{Results obtained in the multiclass setting on different folds in the case with 15 fully-annotated images and 213 weakly-annotated images.}
\begin{tabular}{|c|c|c|c|c|c|c|}
  \hline
 & \scriptsize{fold1}& \scriptsize{fold2}& \scriptsize{fold3}& \scriptsize{fold4}& \scriptsize{fold5}& \scriptsize{mean}\\
\hline
\scriptsize{Standard supervision, whole tumor} &\scriptsize{74.31}&\scriptsize{78.91}&\scriptsize{67.57}&\scriptsize{75.55}&\scriptsize{75.96}&\scriptsize{74.46}\\ 
\hline
\scriptsize{Mixed supervision, whole tumor}&\scriptsize{\textbf{77.53}}&\scriptsize{\textbf{82.20}}&\scriptsize{\textbf{73.72}}&\scriptsize{\textbf{80.96}}&\scriptsize{\textbf{82.55}}&\scriptsize{\textbf{79.39}}\\ 
\hline
\hline
\hline
\scriptsize{Standard supervision, tumor core}&\scriptsize{61.17}&\scriptsize{63.89}&\scriptsize{55.72}&\scriptsize{55.36}&\scriptsize{63.18}&\scriptsize{59.87}\\ 
\hline
\scriptsize{Mixed supervision, tumor core}& \scriptsize{\textbf{62.83}}&\scriptsize{\textbf{65.26}}&\scriptsize{\textbf{62.23}}&\scriptsize{\textbf{61.99}}&\scriptsize{\textbf{67.23}}&\scriptsize{\textbf{63.91}}\\ 
\hline
\hline
\hline
\scriptsize{Standard supervision, enhancing core}&\scriptsize{66.15}&\scriptsize{64.83}&\scriptsize{53.83}&\scriptsize{61.68}&\scriptsize{62.77}&\scriptsize{61.85}\\ 
\hline
\scriptsize{Mixed supervision, enhancing core}& \scriptsize{\textbf{68.33}}&\scriptsize{\textbf{68.39}}&\scriptsize{\textbf{59.51}}&\scriptsize{\textbf{68.63}}&\scriptsize{\textbf{63.70}}&\scriptsize{\textbf{65.71}}\\ 
\hline

  \hline
\end{tabular}
\label{tab_folds2}
\end{table}

In terms of the annotation cost, manual delineation of tumor tissues in one MRI may take about 45 minutes for an experienced oncologist using a dedicated segmentation tool. Determining the range of axial slices containing tumor tissues may take 1-2 minutes but can be done without a specialized software. More importantly, determining global labels may require less medical expertise than performing an exact tumor delineation and can therefore be performed by a larger community.

\section{Conclusion and future work}
In this paper we proposed a new deep learning approach for tumor segmentation which takes advantage of weakly-annotated medical images during the training of neural networks, in addition to a small number of manually segmented images. In our approach, we propose to use neural networks producing both voxelwise and image-level outputs. The classification and segmentation subnetworks share most of their layers and are trained jointly using both fully-annotated and weakly-annotated data. We performed a large number of cross-validated experiments to test our method both in binary and multiclass settings. Our experiments showed that the use of weakly-annotated data improves the segmentation performance significantly when the number of manually segmented images is limited. Our model is end-to-end and straightforward to implement with common deep learning libraries such as Theano \cite{bergstra2010theano} or TensorFlow \cite{abadi2016tensorflow}. The code of our method will be made publicly available in order to encourage other researchers to continue the research in the field.

An interesting step of the future work would be to extend our method to an end-to-end 3D segmentation. In our paper we focused on the 2D segmentation problem, in particular because all 3D images from the BRATS 2018 database contain tumors whereas we also need non-tumor images to train the classification part of our model. One advantage of the 3D application would be to take into account a richer spatial context in the case of MRI or CT scans. Furthermore, volume-level labels require less effort than slice-level labels and would therefore be easier to obtain, even if these labels are also less informative. 

In our tests, we used approximately 220 weakly-annotated MRI, which is relatively a limited number. An important future step would be to test our method on a database containing a considerably larger number of weakly-annotated images (thousands, millions).

\section*{Acknowledgements}
Pawel Mlynarski is funded by the Microsoft Research-INRIA Joint Center, France. This work was supported by the Inria Sophia Antipolis - M\'editerran\'ee, "NEF" computation cluster.









\bibliographystyle{elsarticle-num-names} 
\bibliography{biblio}

\begin{thebibliography}{41}
\expandafter\ifx\csname natexlab\endcsname\relax\def\natexlab#1{#1}\fi
\providecommand{\url}[1]{\texttt{#1}}
\providecommand{\href}[2]{#2}
\providecommand{\path}[1]{#1}
\providecommand{\DOIprefix}{doi:}
\providecommand{\ArXivprefix}{arXiv:}
\providecommand{\URLprefix}{URL: }
\providecommand{\Pubmedprefix}{pmid:}
\providecommand{\doi}[1]{\href{http://dx.doi.org/#1}{\path{#1}}}
\providecommand{\Pubmed}[1]{\href{pmid:#1}{\path{#1}}}
\providecommand{\bibinfo}[2]{#2}
\ifx\xfnm\relax \def\xfnm[#1]{\unskip,\space#1}\fi
\bibitem[{Goodenberger and Jenkins(2012)}]{goodenberger2012genetics}
\bibinfo{author}{M.~L. Goodenberger}, \bibinfo{author}{R.~B. Jenkins},
\newblock \bibinfo{title}{Genetics of adult glioma},
\newblock \bibinfo{journal}{Cancer genetics} \bibinfo{volume}{205}
  (\bibinfo{year}{2012}) \bibinfo{pages}{613--621}.
\bibitem[{Bauer et~al.(2013)Bauer, Wiest, Nolte, and Reyes}]{bauer2013survey}
\bibinfo{author}{S.~Bauer}, \bibinfo{author}{R.~Wiest}, \bibinfo{author}{L.-P.
  Nolte}, \bibinfo{author}{M.~Reyes},
\newblock \bibinfo{title}{A survey of mri-based medical image analysis for
  brain tumor studies},
\newblock \bibinfo{journal}{Physics in medicine and biology}
  \bibinfo{volume}{58} (\bibinfo{year}{2013}) \bibinfo{pages}{R97}.
\bibitem[{LeCun et~al.(1995)LeCun, Bengio et~al.}]{lecun1995convolutional}
\bibinfo{author}{Y.~LeCun}, \bibinfo{author}{Y.~Bengio}, et~al.,
\newblock \bibinfo{title}{Convolutional networks for images, speech, and time
  series},
\newblock \bibinfo{journal}{The handbook of brain theory and neural networks}
  \bibinfo{volume}{3361} (\bibinfo{year}{1995}) \bibinfo{pages}{1995}.
\bibitem[{Long et~al.(2015)Long, Shelhamer, and Darrell}]{long2015fully}
\bibinfo{author}{J.~Long}, \bibinfo{author}{E.~Shelhamer},
  \bibinfo{author}{T.~Darrell},
\newblock \bibinfo{title}{Fully convolutional networks for semantic
  segmentation},
\newblock in: \bibinfo{booktitle}{Proceedings of the IEEE Conference on
  Computer Vision and Pattern Recognition}, \bibinfo{year}{2015}, pp.
  \bibinfo{pages}{3431--3440}.
\bibitem[{Pereira et~al.(2015)Pereira, Pinto, Alves, and
  Silva}]{pereira2015deep}
\bibinfo{author}{S.~Pereira}, \bibinfo{author}{A.~Pinto},
  \bibinfo{author}{V.~Alves}, \bibinfo{author}{C.~A. Silva},
\newblock \bibinfo{title}{Deep convolutional neural networks for the
  segmentation of gliomas in multi-sequence mri},
\newblock in: \bibinfo{booktitle}{International Workshop on Brainlesion:
  Glioma, Multiple Sclerosis, Stroke and Traumatic Brain Injuries},
  \bibinfo{organization}{Springer}, \bibinfo{year}{2015}, pp.
  \bibinfo{pages}{131--143}.
\bibitem[{Kamnitsas et~al.(2016)Kamnitsas, Ledig, Newcombe, Simpson, Kane,
  Menon, Rueckert, and Glocker}]{kamnitsas2016efficient}
\bibinfo{author}{K.~Kamnitsas}, \bibinfo{author}{C.~Ledig},
  \bibinfo{author}{V.~F. Newcombe}, \bibinfo{author}{J.~P. Simpson},
  \bibinfo{author}{A.~D. Kane}, \bibinfo{author}{D.~K. Menon},
  \bibinfo{author}{D.~Rueckert}, \bibinfo{author}{B.~Glocker},
\newblock \bibinfo{title}{Efficient multi-scale 3d cnn with fully connected crf
  for accurate brain lesion segmentation},
\newblock \bibinfo{journal}{arXiv preprint arXiv:1603.05959}
  (\bibinfo{year}{2016}).
\bibitem[{Kamnitsas et~al.(2017)Kamnitsas, Bai, Ferrante, McDonagh, Sinclair,
  Pawlowski, Rajchl, Lee, Kainz, Rueckert et~al.}]{kamnitsas2017ensembles}
\bibinfo{author}{K.~Kamnitsas}, \bibinfo{author}{W.~Bai},
  \bibinfo{author}{E.~Ferrante}, \bibinfo{author}{S.~McDonagh},
  \bibinfo{author}{M.~Sinclair}, \bibinfo{author}{N.~Pawlowski},
  \bibinfo{author}{M.~Rajchl}, \bibinfo{author}{M.~Lee},
  \bibinfo{author}{B.~Kainz}, \bibinfo{author}{D.~Rueckert}, et~al.,
\newblock \bibinfo{title}{Ensembles of multiple models and architectures for
  robust brain tumour segmentation},
\newblock \bibinfo{journal}{arXiv preprint arXiv:1711.01468}
  (\bibinfo{year}{2017}).
\bibitem[{Wang et~al.(2017)Wang, Li, Ourselin, and
  Vercauteren}]{wang2017automatic}
\bibinfo{author}{G.~Wang}, \bibinfo{author}{W.~Li},
  \bibinfo{author}{S.~Ourselin}, \bibinfo{author}{T.~Vercauteren},
\newblock \bibinfo{title}{Automatic brain tumor segmentation using cascaded
  anisotropic convolutional neural networks},
\newblock \bibinfo{journal}{arXiv preprint arXiv:1709.00382}
  (\bibinfo{year}{2017}).
\bibitem[{Menze et~al.(2015)Menze, Jakab, Bauer, Kalpathy-Cramer, Farahani,
  Kirby, Burren, Porz, Slotboom, Wiest et~al.}]{menze2015multimodal}
\bibinfo{author}{B.~H. Menze}, \bibinfo{author}{A.~Jakab},
  \bibinfo{author}{S.~Bauer}, \bibinfo{author}{J.~Kalpathy-Cramer},
  \bibinfo{author}{K.~Farahani}, \bibinfo{author}{J.~Kirby},
  \bibinfo{author}{Y.~Burren}, \bibinfo{author}{N.~Porz},
  \bibinfo{author}{J.~Slotboom}, \bibinfo{author}{R.~Wiest}, et~al.,
\newblock \bibinfo{title}{The multimodal brain tumor image segmentation
  benchmark (brats)},
\newblock \bibinfo{journal}{IEEE transactions on medical imaging}
  \bibinfo{volume}{34} (\bibinfo{year}{2015}) \bibinfo{pages}{1993--2024}.
\bibitem[{Bakas et~al.(2017)Bakas, Akbari, Sotiras, Bilello, Rozycki, Kirby,
  Freymann, Farahani, and Davatzikos}]{bakas2017advancing}
\bibinfo{author}{S.~Bakas}, \bibinfo{author}{H.~Akbari},
  \bibinfo{author}{A.~Sotiras}, \bibinfo{author}{M.~Bilello},
  \bibinfo{author}{M.~Rozycki}, \bibinfo{author}{J.~S. Kirby},
  \bibinfo{author}{J.~B. Freymann}, \bibinfo{author}{K.~Farahani},
  \bibinfo{author}{C.~Davatzikos},
\newblock \bibinfo{title}{Advancing the cancer genome atlas glioma mri
  collections with expert segmentation labels and radiomic features},
\newblock \bibinfo{journal}{Scientific data} \bibinfo{volume}{4}
  (\bibinfo{year}{2017}) \bibinfo{pages}{170117}.
\bibitem[{Ronneberger et~al.(2015)Ronneberger, Fischer, and
  Brox}]{ronneberger2015u}
\bibinfo{author}{O.~Ronneberger}, \bibinfo{author}{P.~Fischer},
  \bibinfo{author}{T.~Brox},
\newblock \bibinfo{title}{U-net: Convolutional networks for biomedical image
  segmentation},
\newblock in: \bibinfo{booktitle}{International Conference on Medical Image
  Computing and Computer-Assisted Intervention},
  \bibinfo{organization}{Springer}, \bibinfo{year}{2015}, pp.
  \bibinfo{pages}{234--241}.
\bibitem[{Pathak et~al.(2014)Pathak, Shelhamer, Long, and
  Darrell}]{pathak2014fully}
\bibinfo{author}{D.~Pathak}, \bibinfo{author}{E.~Shelhamer},
  \bibinfo{author}{J.~Long}, \bibinfo{author}{T.~Darrell},
\newblock \bibinfo{title}{Fully convolutional multi-class multiple instance
  learning},
\newblock \bibinfo{journal}{arXiv preprint arXiv:1412.7144}
  (\bibinfo{year}{2014}).
\bibitem[{Pinheiro and Collobert(2015)}]{pinheiro2015image}
\bibinfo{author}{P.~O. Pinheiro}, \bibinfo{author}{R.~Collobert},
\newblock \bibinfo{title}{From image-level to pixel-level labeling with
  convolutional networks},
\newblock in: \bibinfo{booktitle}{Proceedings of the IEEE Conference on
  Computer Vision and Pattern Recognition}, \bibinfo{year}{2015}, pp.
  \bibinfo{pages}{1713--1721}.
\bibitem[{Saleh et~al.(2016)Saleh, Aliakbarian, Salzmann, Petersson, Gould, and
  Alvarez}]{saleh2016built}
\bibinfo{author}{F.~Saleh}, \bibinfo{author}{M.~S. Aliakbarian},
  \bibinfo{author}{M.~Salzmann}, \bibinfo{author}{L.~Petersson},
  \bibinfo{author}{S.~Gould}, \bibinfo{author}{J.~M. Alvarez},
\newblock \bibinfo{title}{Built-in foreground/background prior for
  weakly-supervised semantic segmentation},
\newblock in: \bibinfo{booktitle}{European Conference on Computer Vision},
  \bibinfo{organization}{Springer}, \bibinfo{year}{2016}, pp.
  \bibinfo{pages}{413--432}.
\bibitem[{Bearman et~al.(2016)Bearman, Russakovsky, Ferrari, and
  Fei-Fei}]{bearman2016s}
\bibinfo{author}{A.~Bearman}, \bibinfo{author}{O.~Russakovsky},
  \bibinfo{author}{V.~Ferrari}, \bibinfo{author}{L.~Fei-Fei},
\newblock \bibinfo{title}{What’s the point: Semantic segmentation with point
  supervision},
\newblock in: \bibinfo{booktitle}{European Conference on Computer Vision},
  \bibinfo{organization}{Springer}, \bibinfo{year}{2016}, pp.
  \bibinfo{pages}{549--565}.
\bibitem[{Wang et~al.(2018)Wang, Liu, Cheng, Wang, Yang, and
  Cheng}]{wang2018automated}
\bibinfo{author}{Z.~Wang}, \bibinfo{author}{C.~Liu},
  \bibinfo{author}{D.~Cheng}, \bibinfo{author}{L.~Wang},
  \bibinfo{author}{X.~Yang}, \bibinfo{author}{K.-T. Cheng},
\newblock \bibinfo{title}{Automated detection of clinically significant
  prostate cancer in mp-mri images based on an end-to-end deep neural network},
\newblock \bibinfo{journal}{IEEE transactions on medical imaging}
  \bibinfo{volume}{37} (\bibinfo{year}{2018}) \bibinfo{pages}{1127--1139}.
\bibitem[{Dreyfus(1990)}]{dreyfus1990artificial}
\bibinfo{author}{S.~E. Dreyfus},
\newblock \bibinfo{title}{Artificial neural networks, back propagation, and the
  kelley-bryson gradient procedure},
\newblock \bibinfo{journal}{Journal of Guidance, Control, and Dynamics}
  \bibinfo{volume}{13} (\bibinfo{year}{1990}) \bibinfo{pages}{926--928}.
\bibitem[{Boyd and Vandenberghe(2004)}]{boyd2004convex}
\bibinfo{author}{S.~Boyd}, \bibinfo{author}{L.~Vandenberghe},
  \bibinfo{title}{Convex optimization}, \bibinfo{publisher}{Cambridge
  university press}, \bibinfo{year}{2004}.
\bibitem[{Girshick et~al.(2014)Girshick, Donahue, Darrell, and
  Malik}]{girshick2014rich}
\bibinfo{author}{R.~Girshick}, \bibinfo{author}{J.~Donahue},
  \bibinfo{author}{T.~Darrell}, \bibinfo{author}{J.~Malik},
\newblock \bibinfo{title}{Rich feature hierarchies for accurate object
  detection and semantic segmentation},
\newblock in: \bibinfo{booktitle}{Proceedings of the IEEE conference on
  computer vision and pattern recognition}, \bibinfo{year}{2014}, pp.
  \bibinfo{pages}{580--587}.
\bibitem[{Oquab et~al.(2015)Oquab, Bottou, Laptev, and Sivic}]{oquab2015object}
\bibinfo{author}{M.~Oquab}, \bibinfo{author}{L.~Bottou},
  \bibinfo{author}{I.~Laptev}, \bibinfo{author}{J.~Sivic},
\newblock \bibinfo{title}{Is object localization for free?-weakly-supervised
  learning with convolutional neural networks},
\newblock in: \bibinfo{booktitle}{Proceedings of the IEEE Conference on
  Computer Vision and Pattern Recognition}, \bibinfo{year}{2015}, pp.
  \bibinfo{pages}{685--694}.
\bibitem[{Simonyan and Zisserman(2014)}]{simonyan2014very}
\bibinfo{author}{K.~Simonyan}, \bibinfo{author}{A.~Zisserman},
\newblock \bibinfo{title}{Very deep convolutional networks for large-scale
  image recognition},
\newblock \bibinfo{journal}{arXiv preprint arXiv:1409.1556}
  (\bibinfo{year}{2014}).
\bibitem[{Krizhevsky et~al.(2012)Krizhevsky, Sutskever, and
  Hinton}]{krizhevsky2012imagenet}
\bibinfo{author}{A.~Krizhevsky}, \bibinfo{author}{I.~Sutskever},
  \bibinfo{author}{G.~E. Hinton},
\newblock \bibinfo{title}{Imagenet classification with deep convolutional
  neural networks},
\newblock in: \bibinfo{booktitle}{Advances in neural information processing
  systems}, \bibinfo{year}{2012}, pp. \bibinfo{pages}{1097--1105}.
\bibitem[{Simonyan et~al.(2013)Simonyan, Vedaldi, and
  Zisserman}]{simonyan2013deep}
\bibinfo{author}{K.~Simonyan}, \bibinfo{author}{A.~Vedaldi},
  \bibinfo{author}{A.~Zisserman},
\newblock \bibinfo{title}{Deep inside convolutional networks: Visualising image
  classification models and saliency maps},
\newblock \bibinfo{journal}{arXiv preprint arXiv:1312.6034}
  (\bibinfo{year}{2013}).
\bibitem[{Bergamo et~al.(2014)Bergamo, Bazzani, Anguelov, and
  Torresani}]{bergamo2014self}
\bibinfo{author}{A.~Bergamo}, \bibinfo{author}{L.~Bazzani},
  \bibinfo{author}{D.~Anguelov}, \bibinfo{author}{L.~Torresani},
\newblock \bibinfo{title}{Self-taught object localization with deep networks},
\newblock \bibinfo{journal}{arXiv preprint arXiv:1409.3964}
  (\bibinfo{year}{2014}).
\bibitem[{Cheplygina et~al.(2018)Cheplygina, de~Bruijne, and
  Pluim}]{cheplygina2018not}
\bibinfo{author}{V.~Cheplygina}, \bibinfo{author}{M.~de~Bruijne},
  \bibinfo{author}{J.~P. Pluim},
\newblock \bibinfo{title}{Not-so-supervised: a survey of semi-supervised,
  multi-instance, and transfer learning in medical image analysis},
\newblock \bibinfo{journal}{arXiv preprint arXiv:1804.06353}
  (\bibinfo{year}{2018}).
\bibitem[{Kamnitsas et~al.(2018)Kamnitsas, Castro, Folgoc, Walker, Tanno,
  Rueckert, Glocker, Criminisi, and Nori}]{kamnitsas2018semi}
\bibinfo{author}{K.~Kamnitsas}, \bibinfo{author}{D.~C. Castro},
  \bibinfo{author}{L.~L. Folgoc}, \bibinfo{author}{I.~Walker},
  \bibinfo{author}{R.~Tanno}, \bibinfo{author}{D.~Rueckert},
  \bibinfo{author}{B.~Glocker}, \bibinfo{author}{A.~Criminisi},
  \bibinfo{author}{A.~Nori},
\newblock \bibinfo{title}{Semi-supervised learning via compact latent space
  clustering},
\newblock \bibinfo{journal}{arXiv preprint arXiv:1806.02679}
  (\bibinfo{year}{2018}).
\bibitem[{Zhang et~al.(2001)Zhang, Brady, and Smith}]{zhang2001segmentation}
\bibinfo{author}{Y.~Zhang}, \bibinfo{author}{M.~Brady},
  \bibinfo{author}{S.~Smith},
\newblock \bibinfo{title}{Segmentation of brain mr images through a hidden
  markov random field model and the expectation-maximization algorithm},
\newblock \bibinfo{journal}{IEEE transactions on medical imaging}
  \bibinfo{volume}{20} (\bibinfo{year}{2001}) \bibinfo{pages}{45--57}.
\bibitem[{Papandreou et~al.(2015)Papandreou, Chen, Murphy, and
  Yuille}]{papandreou2015weakly}
\bibinfo{author}{G.~Papandreou}, \bibinfo{author}{L.-C. Chen},
  \bibinfo{author}{K.~P. Murphy}, \bibinfo{author}{A.~L. Yuille},
\newblock \bibinfo{title}{Weakly-and semi-supervised learning of a deep
  convolutional network for semantic image segmentation},
\newblock in: \bibinfo{booktitle}{Proceedings of the IEEE international
  conference on computer vision}, \bibinfo{year}{2015}, pp.
  \bibinfo{pages}{1742--1750}.
\bibitem[{Rajchl et~al.(2016)Rajchl, Lee, Oktay, Kamnitsas, Passerat-Palmbach,
  Bai, Kainz, and Rueckert}]{rajchl2016deepcut}
\bibinfo{author}{M.~Rajchl}, \bibinfo{author}{M.~C. Lee},
  \bibinfo{author}{O.~Oktay}, \bibinfo{author}{K.~Kamnitsas},
  \bibinfo{author}{J.~Passerat-Palmbach}, \bibinfo{author}{W.~Bai},
  \bibinfo{author}{B.~Kainz}, \bibinfo{author}{D.~Rueckert},
\newblock \bibinfo{title}{Deepcut: Object segmentation from bounding box
  annotations using convolutional neural networks},
\newblock \bibinfo{journal}{arXiv preprint arXiv:1605.07866}
  (\bibinfo{year}{2016}).
\bibitem[{Hung et~al.(2018)Hung, Tsai, Liou, Lin, and
  Yang}]{hung2018adversarial}
\bibinfo{author}{W.-C. Hung}, \bibinfo{author}{Y.-H. Tsai},
  \bibinfo{author}{Y.-T. Liou}, \bibinfo{author}{Y.-Y. Lin},
  \bibinfo{author}{M.-H. Yang},
\newblock \bibinfo{title}{Adversarial learning for semi-supervised semantic
  segmentation},
\newblock \bibinfo{journal}{arXiv preprint arXiv:1802.07934}
  (\bibinfo{year}{2018}).
\bibitem[{Goodfellow et~al.(2014)Goodfellow, Pouget-Abadie, Mirza, Xu,
  Warde-Farley, Ozair, Courville, and Bengio}]{goodfellow2014generative}
\bibinfo{author}{I.~Goodfellow}, \bibinfo{author}{J.~Pouget-Abadie},
  \bibinfo{author}{M.~Mirza}, \bibinfo{author}{B.~Xu},
  \bibinfo{author}{D.~Warde-Farley}, \bibinfo{author}{S.~Ozair},
  \bibinfo{author}{A.~Courville}, \bibinfo{author}{Y.~Bengio},
\newblock \bibinfo{title}{Generative adversarial nets},
\newblock in: \bibinfo{booktitle}{Advances in neural information processing
  systems}, \bibinfo{year}{2014}, pp. \bibinfo{pages}{2672--2680}.
\bibitem[{Hong et~al.(2015)Hong, Noh, and Han}]{hong2015decoupled}
\bibinfo{author}{S.~Hong}, \bibinfo{author}{H.~Noh}, \bibinfo{author}{B.~Han},
\newblock \bibinfo{title}{Decoupled deep neural network for semi-supervised
  semantic segmentation},
\newblock in: \bibinfo{booktitle}{Advances in neural information processing
  systems}, \bibinfo{year}{2015}, pp. \bibinfo{pages}{1495--1503}.
\bibitem[{Hong et~al.(2016)Hong, Oh, Lee, and Han}]{hong2016learning}
\bibinfo{author}{S.~Hong}, \bibinfo{author}{J.~Oh}, \bibinfo{author}{H.~Lee},
  \bibinfo{author}{B.~Han},
\newblock \bibinfo{title}{Learning transferrable knowledge for semantic
  segmentation with deep convolutional neural network},
\newblock in: \bibinfo{booktitle}{Proceedings of the IEEE Conference on
  Computer Vision and Pattern Recognition}, \bibinfo{year}{2016}, pp.
  \bibinfo{pages}{3204--3212}.
\bibitem[{Evgeniou and Pontil(2004)}]{evgeniou2004regularized}
\bibinfo{author}{T.~Evgeniou}, \bibinfo{author}{M.~Pontil},
\newblock \bibinfo{title}{Regularized multi--task learning},
\newblock in: \bibinfo{booktitle}{Proceedings of the tenth ACM SIGKDD
  international conference on Knowledge discovery and data mining},
  \bibinfo{organization}{ACM}, \bibinfo{year}{2004}, pp.
  \bibinfo{pages}{109--117}.
\bibitem[{Shah et~al.(2018)Shah, Merchant, and Awate}]{shah2018ms}
\bibinfo{author}{M.~P. Shah}, \bibinfo{author}{S.~Merchant},
  \bibinfo{author}{S.~P. Awate},
\newblock \bibinfo{title}{Ms-net: Mixed-supervision fully-convolutional
  networks for full-resolution segmentation},
\newblock in: \bibinfo{booktitle}{International Conference on Medical Image
  Computing and Computer-Assisted Intervention},
  \bibinfo{organization}{Springer}, \bibinfo{year}{2018}, pp.
  \bibinfo{pages}{379--387}.
\bibitem[{Ioffe and Szegedy(2015)}]{ioffe2015batch}
\bibinfo{author}{S.~Ioffe}, \bibinfo{author}{C.~Szegedy},
\newblock \bibinfo{title}{Batch normalization: Accelerating deep network
  training by reducing internal covariate shift},
\newblock \bibinfo{journal}{arXiv preprint arXiv:1502.03167}
  (\bibinfo{year}{2015}).
\bibitem[{He et~al.(2016)He, Zhang, Ren, and Sun}]{he2016deep}
\bibinfo{author}{K.~He}, \bibinfo{author}{X.~Zhang}, \bibinfo{author}{S.~Ren},
  \bibinfo{author}{J.~Sun},
\newblock \bibinfo{title}{Deep residual learning for image recognition},
\newblock in: \bibinfo{booktitle}{Proceedings of the IEEE conference on
  computer vision and pattern recognition}, \bibinfo{year}{2016}, pp.
  \bibinfo{pages}{770--778}.
\bibitem[{Mlynarski et~al.(2018)Mlynarski, Delingette, Criminisi, and
  Ayache}]{mlynarski20183d}
\bibinfo{author}{P.~Mlynarski}, \bibinfo{author}{H.~Delingette},
  \bibinfo{author}{A.~Criminisi}, \bibinfo{author}{N.~Ayache},
\newblock \bibinfo{title}{3d convolutional neural networks for tumor
  segmentation using long-range 2d context},
\newblock \bibinfo{journal}{arXiv preprint arXiv:1807.08599}
  (\bibinfo{year}{2018}).
\bibitem[{Rumelhart et~al.(1988)Rumelhart, Hinton, and
  Williams}]{rumelhart1988learning}
\bibinfo{author}{D.~E. Rumelhart}, \bibinfo{author}{G.~E. Hinton},
  \bibinfo{author}{R.~J. Williams},
\newblock \bibinfo{title}{Learning representations by back-propagating errors},
\newblock \bibinfo{journal}{Cognitive modeling} \bibinfo{volume}{5}
  (\bibinfo{year}{1988}) \bibinfo{pages}{1}.
\bibitem[{Bergstra et~al.(2010)Bergstra, Breuleux, Bastien, Lamblin, Pascanu,
  Desjardins, Turian, Warde-Farley, and Bengio}]{bergstra2010theano}
\bibinfo{author}{J.~Bergstra}, \bibinfo{author}{O.~Breuleux},
  \bibinfo{author}{F.~Bastien}, \bibinfo{author}{P.~Lamblin},
  \bibinfo{author}{R.~Pascanu}, \bibinfo{author}{G.~Desjardins},
  \bibinfo{author}{J.~Turian}, \bibinfo{author}{D.~Warde-Farley},
  \bibinfo{author}{Y.~Bengio},
\newblock \bibinfo{title}{Theano: A cpu and gpu math compiler in python},
\newblock in: \bibinfo{booktitle}{Proc. 9th Python in Science Conf},
  \bibinfo{year}{2010}, pp. \bibinfo{pages}{1--7}.
\bibitem[{Abadi et~al.(2016)Abadi, Agarwal, Barham, Brevdo, Chen, Citro,
  Corrado, Davis, Dean, Devin et~al.}]{abadi2016tensorflow}
\bibinfo{author}{M.~Abadi}, \bibinfo{author}{A.~Agarwal},
  \bibinfo{author}{P.~Barham}, \bibinfo{author}{E.~Brevdo},
  \bibinfo{author}{Z.~Chen}, \bibinfo{author}{C.~Citro}, \bibinfo{author}{G.~S.
  Corrado}, \bibinfo{author}{A.~Davis}, \bibinfo{author}{J.~Dean},
  \bibinfo{author}{M.~Devin}, et~al.,
\newblock \bibinfo{title}{Tensorflow: Large-scale machine learning on
  heterogeneous distributed systems},
\newblock \bibinfo{journal}{arXiv preprint arXiv:1603.04467}
  (\bibinfo{year}{2016}).

\end{thebibliography}

\end{document}